\pdfoutput=1

\documentclass[11pt]{article}

\usepackage[preprint]{acl}

\usepackage{times}
\usepackage{latexsym}
\usepackage{amsmath}
\usepackage{amsthm}

\usepackage[T1]{fontenc}

\usepackage[utf8]{inputenc}

\usepackage{microtype}

\usepackage{inconsolata}

\usepackage{graphicx}

\title{HalluGuard: Evidence-Grounded Small Reasoning Models to Mitigate Hallucinations in Retrieval-Augmented Generation}

\author{
\textbf{Loris Bergeron}\textsuperscript{1,4} \quad
\textbf{Ioana Buhnila}\textsuperscript{2,3} \quad
\textbf{Jérôme François}\textsuperscript{4} \quad
\textbf{Radu State}\textsuperscript{4} \\
\textsuperscript{1}Banque de Luxembourg \quad
\textsuperscript{2}Center for Data Science in Humanities, Chosun University \\
\textsuperscript{3}ATILF, University of Lorraine--CNRS \quad
\textsuperscript{4}SnT, University of Luxembourg \\
\small{\textbf{Correspondence:} \texttt{loris.bergeron@blu.bank}}
}

\usepackage{booktabs}
\usepackage{makecell}
\usepackage{multirow}
\usepackage{colortbl}
\usepackage{xcolor}
\usepackage{float}

\usepackage{pgfplots}
\usepackage{pgfplotstable}
\usetikzlibrary{arrows}
\pgfplotsset{compat=1.18}
\usepgfplotslibrary{polar}
\usetikzlibrary{patterns}
\usetikzlibrary{calc}

\usepackage{tabularx}
\usepackage{booktabs}

\definecolor{verifycolor}{RGB}{220,20,60}
\definecolor{okcolor}{RGB}{0,100,0}

\definecolor{darkgreen}{RGB}{0,169,137}
\definecolor{lightgreen}{RGB}{195,253,158}

\usepackage{textcomp}
\usepackage{listings}
\lstdefinestyle{acljson}{
  basicstyle=\ttfamily\small,
  columns=fullflexible,
  keepspaces=true,
  showstringspaces=false,
  breaklines=true,
  breakatwhitespace=true,
  breakautoindent=true,
  breakindent=2em,
  frame=none,
  xleftmargin=0pt,
  upquote=true
}

\usepackage{enumitem}

\newcommand{\CG}{\textsc{CG}}
\newcommand{\DR}{\textsc{DR}}
\newcommand{\PGL}{\textsc{PG-L}}
\newcommand{\PGS}{\textsc{PG-S}}
\newcommand{\IEone}{\textsc{IE-1}}
\newcommand{\IEtwo}{\textsc{IE-2}}

\newcommand{\QwenThirtyTwo}{Qwen3-32B}
\newcommand{\QwenZeroSix}{Qwen3-0.6B}
\newcommand{\QwenFourB}{Qwen3-4B}
\newcommand{\LlamaSeventy}{Llama-3.3-70B}
\newcommand{\MistralLtwo}{Mistral Large 2}
\newcommand{\HalluGuard}{HalluGuard-4B}

\begin{document}
\maketitle

\begin{abstract}
Large Language Models (LLMs) excel in many NLP tasks but remain prone to hallucinations, limiting trust in real-world applications. We present HalluGuard, a 4B-parameter Small Reasoning Model (SRM) for mitigating hallucinations in Retrieval-Augmented Generation (RAG). HalluGuard classifies document–claim pairs as grounded or hallucinated and produces evidence-grounded justifications for transparency. Our approach combines (i) a domain-agnostic synthetic dataset derived from FineWeb and refined through multi-stage curation and data reformation, (ii) synthetic grounded and hallucinated claims, and (iii) preference-based fine-tuning with Odds Ratio Preference Optimization to distill large-model reasoning into a smaller backbone. On the RAGTruth subset of the LLM-AggreFact benchmark, HalluGuard achieves 84.0\% balanced accuracy (BAcc), rivaling specialized models, MiniCheck (7B; 84.0\%) and Granite Guardian 3.3 (8B; 82.2\%) while using roughly half their parameters. Over the full benchmark it reaches 75.7\% BAcc, matching larger general-purpose LLMs such as GPT-4o (75.9\%). We will release HalluGuard and datasets under Apache 2.0 upon acceptance.
\end{abstract}

\section{Introduction}

Large Language Models (LLMs) have been used for a variety of Natural Language Processing (NLP) tasks, achieving strong results in summarization, text classification, and question answering~\cite{tan2023can, singhal2023large}. 

However, recent research shows that Small Language Models (SLMs)~\cite{schick2021s} can achieve competitive results in specific tasks, especially when fine-tuned on domain-specific data. In addition to being cost and energy efficient, SLMs are practical in resource-constrained settings~\cite{lepagnol2024small} such as on-premise environments, often required in the financial sector and industries with strict compliance requirements.

However, a major remaining challenge is that both LLMs and SLMs are prone to hallucinations, outputs inconsistent with the input prompt or factual knowledge~\cite{zhang2025siren, huang2023survey}, and are problematic in Retrieval-Augmented Generation (RAG)~\cite{lewis2020retrieval} applications, increasingly deployed in companies due to their ability to deliver context-aware responses. 

Even when using documents, RAGs remain vulnerable to hallucinations~\cite{niu-etal-2024-ragtruth}, undermining trust and explainability~\cite{ni2025towards}. To address this, models must be able to detect hallucinations, justify their outputs with evidence, and integrate into RAG applications.

\begin{figure}[t]
\centerline{\includegraphics[width=1\linewidth]{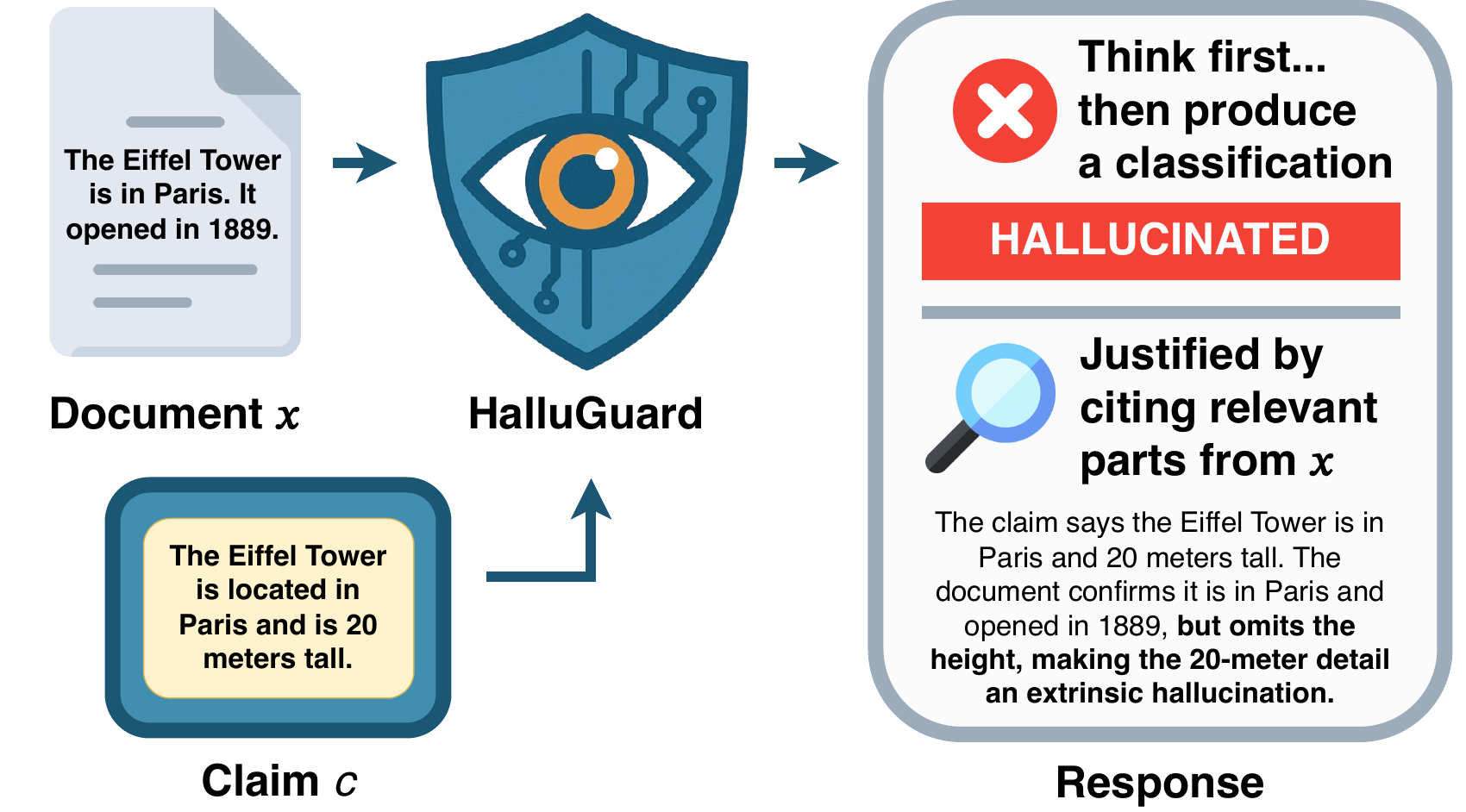}}
\caption{\textbf{HalluGuard Concept.} Given a document $x$ and a claim $c$, the model first thinks before classifying their relationship as \texttt{grounded} or \texttt{hallucinated}, and then produces a justification citing relevant parts of $x$.}
\label{fig:halluguard_concept}
\end{figure}

Recent work emphasizes models designed for reasoning, the ability to perform multi-step inference, follow logical chains, and provide transparent reasoning traces~\cite{wei2022chain}. Small Reasoning Models (SRMs) are not merely SLMs run with Chain-of-Thought (CoT) prompts. Rather, they are trained to produce structured intermediate reasoning that decomposes complex tasks before generating output, often through distillation from stronger reasoners and reward-guided training~\cite{xu2025srm}. This makes SRMs particularly well suited for mitigating hallucinations in RAG applications.

Moreover, most of the previous work on hallucination detection uses BERT-based classifiers~\cite{devlin2019bert}. Although effective, these models do not provide justifications, making them unsuitable when explainability is mandatory.

This challenge is especially pressing in business environments, where regulations require decision-traceable justifications. 
In fact, to improve efficiency, companies, especially in finance, are deploying custom RAG solutions with team-specific knowledge (e.g., compliance, legal). Similar deployments are spreading to other industries where specialized knowledge is critical. In these settings, trust and explainability are essential. Users must see which passages of the retrieved document support or contradict the claim.

To address this gap, we propose HalluGuard, an SRM for the mitigation of hallucinations in RAG. As shown in Figure~\ref{fig:halluguard_concept}, given a document $x$ and claim $c$, HalluGuard first thinks about their relationship before predicting whether the claim is \texttt{grounded} or \texttt{hallucinated}, while generating an evidence-grounded justification, fostering the transparent and reliable use of RAG in companies.

Our contributions are threefold:
\begin{itemize}
    \item We introduce HalluGuard\footnote{\url{https://anonymous.website}}, a Small Reasoning Model (SRM) for hallucination mitigation in Retrieval-Augmented Generation (RAG). HalluGuard detects hallucinations and generates evidence-grounded justifications, making it transparent for human oversight. We will publicly release HalluGuard and the datasets used for fine-tuning upon acceptance.
    
    \item We construct HalluClaim\footnote{\url{https://anonymous.website}}, a large-scale synthetic dataset derived from FineWeb~\cite{penedo2024fineweb} using Llama3.3-70B~\cite{dubey2024llama}. HalluClaim provides a controlled yet diverse benchmark for training and evaluating hallucination detection in RAG-like scenarios and will also be released.
    
    \item We show that HalluGuard improves the balanced accuracy of its backbone and achieves competitive performance compared to larger open-source and closed LLMs. Our ablation study highlights the role of reasoning traces, consensus filtering, and Odds Ratio Preference Optimization (ORPO)~\cite{hong-etal-2024-orpo} fine-tuning in driving these gains.
\end{itemize}

\section{Related Work}

Mitigating hallucinations in LLMs has been approached through prompt engineering, Retrieval-Augmented Generation, decoding strategies, supervised fine-tuning, and self-reflection (\citealt{ji2023towards}; \citealt{song2024rag}; \citealt{tonmoy2024comprehensive}; \citealt{zhang2025llm}). Despite the extensive study of hallucinations in LLMs, there is no consensus on a general classification, as the boundary between hallucination and factuality is often blurred (\citealt{wei2024long}; \citealt{mallen2023not}). To address this, \citeauthor{bang-etal-2025-hallulens} (\citeyear{bang-etal-2025-hallulens}) proposed a three-type taxonomy.

Linked to our work, LYNX~\cite{ravi2024lynx} is an open-source hallucination evaluation model that outperforms GPT-4o and Claude-3 Sonnet, with 8B and 70B-parameter variants. In addition, IBM's Granite Guardian 3.3~\cite{padhi2024graniteguardian}, an 8B model, detects hallucinations in RAG settings and provides yes/no scores with optional reasoning traces through hybrid thinking modes.

Fact-checking has been studied beyond hallucination detection. \citeauthor{tang-etal-2024-minicheck} (\citeyear{tang-etal-2024-minicheck}) introduced MiniCheck, a model trained on synthetic data matching GPT-4 on multi-fact reasoning benchmarks, while remaining more cost-effective. This line of research highlights the importance of lightweight and scalable models for this specific task. 
More recently, \citeauthor{pandit2025teaching} (\citeyear{pandit2025teaching}) presented HaluCheck, a hallucination detection model trained with a curriculum-based Direct Preference Optimization (DPO)~\cite{rafailov2023direct} framework.
HaluCheck was not publicly available at that time.

\section{Problem Formulation} \label{s:problem_formulation}

\begin{figure*}[t]
\centering
\includegraphics[width=0.9\linewidth]{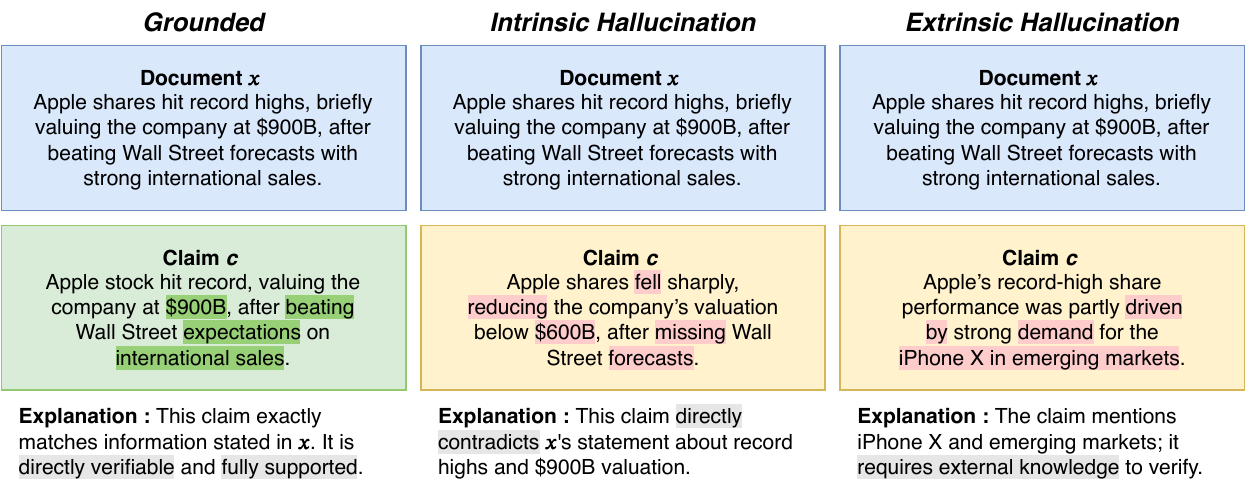}
\caption{\textbf{Examples of Relations.} A grounded claim, an intrinsic hallucination, and an extrinsic hallucination.}
\label{fig:label_categories}
\end{figure*}

We define the task as determining the relationship between a document \(x\) and a claim \(c\).
The relation $t(x, c)$ can take one of three values:
\[ t(x, c) = \begin{cases} \text{grounded}, & \!\!\!\! \text{if $c$ is supported by $x$} \\ \text{intrinsic\_hallu}, & \!\!\!\! \text{if $c$ contradicts $x$} \\ \text{extrinsic\_hallu}, & \!\!\!\! \text{if $c$ is not in $x$} \end{cases} \]

A claim $c$ is grounded if it is fully supported by the information explicitly present in $x$.
It is an intrinsic hallucination if it directly contradicts $x$, and an extrinsic hallucination if its truth requires external knowledge beyond $x$.
Concrete examples of these relationships can be found in Figure~\ref{fig:label_categories}.

For the remainder of this work, we group intrinsic and extrinsic hallucinations under a single \texttt{hallucinated} label, reducing the task to binary classification (\texttt{grounded} vs.\ \texttt{hallucinated}).

\section{Method}
\label{s:method}

\subsection{HalluGuard Overview}
HalluGuard is a Small Reasoning Model (SRM) designed to mitigate hallucinations in Retrieval-Augmented Generation (RAG). Given a document–claim pair, it predicts whether the claim is \texttt{grounded} or \texttt{hallucinated}, and provides a justification citing the document, improving transparency and user trust. HalluGuard supports two inference modes: in the \texttt{think} mode, it generates intermediate reasoning traces before the final output, while in \texttt{non-think} mode, it skips these traces and outputs directly. The mode is controlled at inference time by adding \texttt{/think} or \texttt{/no\_think} to the prompt.

\begin{figure*}[htbp]
\centering
\includegraphics[width=0.80\linewidth]{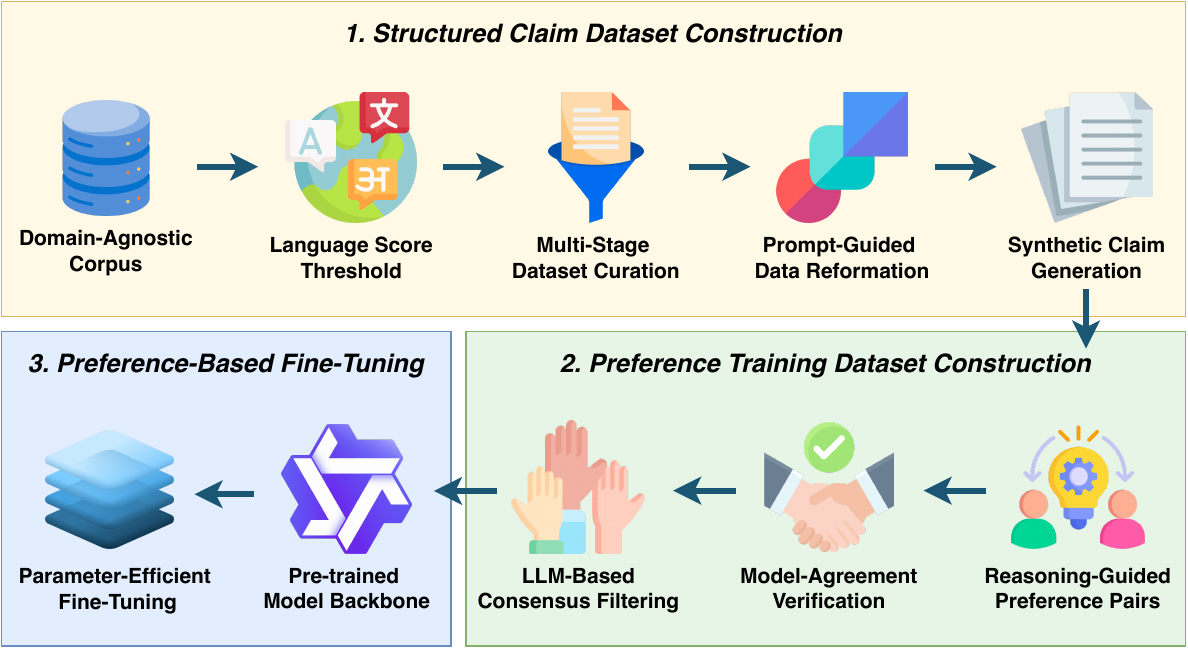}
\caption{\textbf{HalluGuard Training Pipeline.} A domain-agnostic corpus is filtered, reformed, and used to generate three types of synthetic claims (grounded, intrinsic hallucinated, and extrinsic hallucinated). Preference data are built via cross-model generation (Qwen3-32B and Qwen3-0.6B), model-agreement verification and LLM-based consensus filtering are used to enhance quality and confidence. The Qwen3-4B backbone is then fine-tuned using LoRA and ORPO to mitigate hallucinations and produce evidence-grounded justifications in RAG applications.}
\label{fig:halluguard_pipeline}
\end{figure*}

As shown in Figure~\ref{fig:halluguard_pipeline}, our method begins with a large, high-quality, domain-agnostic corpus that has been curated for safety, quality, and diversity. The texts in this corpus are then linguistically reformed in tone and style by the Data Reformer (\DR; \LlamaSeventy{}) to improve cross-domain generalization. From these reformed texts, we generate grounded and hallucinated synthetic claims using the Claim Generator (\CG; \LlamaSeventy{}).

To align the model towards high-quality reasoning and justifications, we construct a synthetic preference dataset. For each document–claim pair, we generate two candidate completions: one from the Preference Generator-Large (\PGL; \QwenThirtyTwo{}) and one from the Preference Generator-Small (\PGS; \QwenZeroSix{}~\cite{qwen3}). We designate the output of the \PGL{} as the \texttt{chosen} response and the output of the \PGS{} as the \texttt{rejected} response. This creates preference pairs that exploit the empirical quality gap between large and small models, enabling us to build a training dataset without the need for additional human annotation. To further improve reliability, we apply two filtering steps: (i) model-agreement verification, in which the label deduced from the synthetic claim (from \CG{}) is compared with the classification produced by \PGL{}; and (ii) LLM-based consensus filtering. In this step, two Independent Evaluators (\IEone{}; \LlamaSeventy{} and \IEtwo{}; \MistralLtwo{}~\cite{mistral_large_2407_2024}) judge both completions. Only pairs in which both evaluators select the \texttt{chosen} completion are retained. Finally, we fine-tune a \QwenFourB{} backbone using LoRA~\cite{hu2022lora} for efficiency and Odds Ratio Preference Optimization (ORPO), which merges Supervised Fine-Tuning (SFT) and preference alignment into a single stage. \QwenFourB{} was selected to avoid the Learnability Gap observed in SLMs~\cite{li-etal-2025-small-models}.

Thus, HalluGuard is a Small Reasoning Model that delivers reliable hallucination mitigation and interpretable justifications, ready for seamless integration into enterprise RAG applications.

\subsection{Structured Claim Dataset Construction}
\label{ss:preparing_the_foundations}

\paragraph{Domain-Agnostic Corpus.} The performance of LLMs depends on both the size and the quality of the dataset~\cite{gunasekar2023textbooksneed}. Larger and more diverse datasets improve generalization by exposing models to varied contexts. We therefore use FineWeb~\cite{penedo2024fineweb}, a large-scale, open-source, domain-agnostic web corpus.

From the 10TB FineWeb sample\footnote{\url{https://hf.co/datasets/HuggingFaceFW/fineweb}}, we retain only documents with a high confidence of being in English (\texttt{language\_score} $\geq$ 0.95) and remove exact duplicates. From the remaining pool, we randomly sample 250{,}000 documents to form the baseline dataset, denoted \(D_{\text{agnostic}}\).

\paragraph{Multi-Stage Dataset Curation.} We further filter \(D_{\text{agnostic}}\) to ensure safety, quality and diversity, following practices similar to C4~\cite{raffel2020exploring}. Without this step, models risk learning unsafe, low-quality, or repetitive patterns. Specifically, we remove documents containing unsafe terms\footnote{\url{https://github.com/LDNOOBW}}, discard those that do not comply with C4-style quality rules (e.g., pages with fewer than five sentences, lines missing terminal punctuation, boilerplate such as \texttt{Lorem Ipsum} or cookie notices, and malformed text such as a single token over 1000 characters). Finally, we remove documents shorter than 50 words and near-duplicates of any three consecutive sentences from documents that have already been retained. This deduplication step is important for promoting diversity: by eliminating redundant content, it reduces the repeated boilerplate and ensures that a wider range of topics and writing styles are represented. The resulting dataset, denoted \(D_{\text{clean}}\), contains 86{,}024 documents.

\paragraph{Prompt-Guided Data Reformation.} Despite multi-stage curation, \( D_\text{clean} \) remains web-centric in style due to the nature of FineWeb. To increase linguistic diversity and improve generalization to non-web formats (e.g., reports, dialogues), we use \DR{} to rewrite each document, producing a wider range of styles that better reflect real-world variation~\cite{veselovsky_generating_2023, long_llms-driven_2024}.

The reformed dataset is then:
\begin{equation}
D_{\text{reformed}} =
\left\{\, s_{j(x)}\big(x; T(x)\big) \;\middle|\; x \in D_{\text{clean}} \,\right\}
\end{equation}
where $j(x)$ is a random style from $\mathcal{S} = \{s_1, s_2, \dots, s_{18}\}$ defined in Appendix~\ref{app:prompt_data_augmentation}, and $T(x)$ the temperature sampled uniformly from $[0.2, 0.7]$.

\paragraph{Synthetic Claim Generation.} We generate one synthetic claim per document in \(D_\text{reformed}\). 
To balance the binary classification task, we generate half \texttt{grounded} and half \texttt{hallucinated} claims, with the \texttt{hallucinated} split evenly into intrinsic and extrinsic, but both labeled as \texttt{hallucinated}.

For each document \(x_i \in D_\text{reformed}\), we ask \CG{} to generate a claim \(c_i\) in structured JSON format~\cite{he2024does} (see Appendix~\ref{app:prompt_claims_generation}), and assign it a label \(t_i \in \{\texttt{grounded}, \texttt{hallucinated}\}\). This results in 86{,}024 balanced document--claim--label triplets:
\begin{equation}
HalluClaim =
\bigcup_{t \in \mathcal{C}} \{(x_i, c_i, t_i) \mid x_i \in D_t\}
\end{equation}

\subsection{Preference Training Dataset Construction}
\label{ss:preference_training_dataset_construction}

\paragraph{Reasoning-Guided Preference Pairs.} The balanced dataset \( HalluClaim\) contains document--claim--label triplets. However, our goal is not only to classify claims correctly, but also to train models to produce evidence-grounded justification.

We convert \( HalluClaim\) into the preference dataset format\footnote{\url{https://hf.co/docs/trl/dataset_formats}}, where each instance comprises a prompt and two completions: a \texttt{chosen} completion and a \texttt{rejected} one (see Appendix~\ref{app:preference_data}). For each triplet of documents--claim--label, we construct a prompt \( P_i \) containing: (i) task instructions defining the \texttt{grounded} and \texttt{hallucinated} labels, (ii) the document \( x_i \) and (iii) the claim \( c_i \). The prompt requires classification and justification (see Appendix~\ref{app:prompt_pairs_generation}).

Thus, we used \PGL{} and \PGS{}, with the same prompt \( P_i \). Each model \( m \in \{\PGL{},\ \PGS{}\} \) produces the response as follows:
\begin{equation}
R_i^{(m)} = \big( y_i^{(m)},\ j_i^{(m)},\ r_i^{(m)} \big)    
\end{equation} where \( y_i^{(m)} \) is the predicted label (\texttt{grounded} or \texttt{hallucinated}), \( j_i^{(m)} \) is the justification and \( r_i^{(m)} \) is the model reasoning within the \texttt{<think>} tags.

Assuming that larger models perform better, we apply a size-based heuristic, marking $R_i^{(\PGL{})}$ as \texttt{chosen} and $R_i^{(\PGS{})}$ as \texttt{rejected}.

For each triplet \((x_i, c_i, t_i)\) in \( HalluClaim\), we produce preference tuples of the form:
\begin{equation}
    z_i = \big( P_i,\ R_i^{(\PGL{})} \ \text{(\scriptsize{chosen})},\ R_i^{(\PGS{})} \ \text{(\scriptsize{rejected})} \big)
\end{equation}

\paragraph{Model-Agreement Verification.} The size-based heuristic provides a useful starting point, but some \texttt{chosen} completions may still misclassify the claim. To correct this, we require agreement between the synthetic label assigned by \CG{} during claim generation and the classification predicted by \PGL{}. Any tuple where the \texttt{chosen} label disagrees with the synthetic label is removed. After this verification, \(HalluClaim_{\text{pref}}\) contains 83{,}020 tuples.

\paragraph{LLM-Based Consensus Filtering.} To further improve reliability, each tuple is independently evaluated by \IEone{} and \IEtwo{} in a few-shot setting~\cite{brown2020language} using a dedicated prompt that asks for the selection of the best completion according to three criteria: (i) classification correctness, (ii) coherence of reasoning, and (iii) clarity of justification (see Appendix~\ref{app:prompt_filtering_generation}).
The models receive the full prompt \(P_i\) and completions, without being told which one is the \texttt{chosen} completion.

A tuple is retained only if \IEone{} and \IEtwo{} select the same completion that matches the \texttt{chosen} one.
\begin{equation}
    \text{\IEone}(P_i) = \text{\IEtwo}(P_i) = R_i^{(\mathrm{chosen})}
\end{equation}

This LLM-based consensus step reduces label noise and mitigates size-based heuristic bias, thereby yielding a total of 75,360 high-quality preference tuples for fine-tuning.

\subsection{Preference-Based Fine-Tuning}
\label{ss:fine_tuning}

\paragraph{Pre-trained Model Backbone.} After creating a high-quality preference dataset through model-agreement verification and consensus filtering, we use \QwenFourB{}~\cite{qwen3} as backbone for fine-tuning.
It supports a context window of up to 32,768 tokens, which is important for document-level reasoning. The 4B variant remains lightweight enough for enterprise on-prem deployment. Using \QwenFourB{}, we address the Small Model Learnability Gap~\cite{li-etal-2025-small-models} observed in models with at most 3B parameters.

\paragraph{Parameter-Efficient Fine-Tuning.} We fine-tune \QwenFourB{} using ORPO, a fine-tuning technique that increases the gap between \texttt{chosen} and \texttt{rejected} completions so that the model consistently favors the \texttt{chosen} one (see Appendix~\ref{app:epoch_reward_margins}). Unlike DPO, ORPO performs an SFT stage during preference alignment, without relying on a reference model. This makes training more efficient and allows HalluGuard to accurately classify claims while generating justifications and reasoning distilled from stronger models.

To apply ORPO in a parameter-efficient manner, we use LoRA~\cite{hu2022lora}, which freezes most base weights and trains only small adapter layers. This reduces memory and compute costs while mitigating catastrophic forgetting when adapting pre-trained models to specific tasks~\cite{bafghi2025fine}. Given the 32k token context window, fine-tuning is memory intensive. We therefore use Unsloth\footnote{\url{https://unsloth.ai}}, which accelerates fine-tuning with custom kernels, and memory optimizations, to enable faster, stable training. Reproducibility and fine-tuning details are in Appendices~\ref{app:reproducibility} and \ref{app:fine_tuning_protocol}.

\begin{table*}[h]
\centering
\small
\resizebox{\textwidth}{!}{%
\begin{tabular}{llcccccccccccccc}
\toprule
\multirow{2}{*}{\textbf{Model}} 
& \multirow{2}{*}{\textbf{Size}} 
& \multicolumn{2}{c}{\textbf{AGGREFACT}} 
& \multicolumn{2}{c}{\textbf{TofuEval}} 
& \multirow{2}{*}{\textbf{WiCE}} 
& \multirow{2}{*}{\textbf{REVEAL}} 
& \multirow{2}{*}{\makecell{\textbf{Claim}\\\textbf{Verify}}} 
& \multirow{2}{*}{\makecell{\textbf{Fact}\\\textbf{Check}}} 
& \multirow{2}{*}{\makecell{\textbf{Expert}\\\textbf{QA}}} 
& \multirow{2}{*}{\textbf{LFQA}} 
& \multirow{2}{*}{\makecell{\textbf{RAG}\\\textbf{Truth}}} 
& \multirow{2}{*}{\makecell{\textbf{BAcc}\\\textbf{Avg.}}} 
\\
& & CNN & XSum & MediaS & MeetB \\
\midrule
Qwen3-32B$^{\star}$ & 32B & 69.1 & 76.3 & 72.0 & 82.2 & 80.6 & 90.0 & 73.3 & 77.9 & 60.2 & 85.5 & 85.9 & 77.6 \\
MiniCheck-7B & 7B & 65.5 & 77.8 & 76.0 & 78.3 & 83.0 & 88.0 & 75.3 & 77.7 & 59.2 & 86.7 & 84.0 & 77.4 \\
Claude-3.5 Sonnet & - & 67.6 & 75.1 & 73.4 & 84.6 & 77.7 & 89.1 & 71.4 & 77.8 & 60.9 & 85.6 & 86.1 & 77.2 \\
Granite Guardian 3.3 & 8B & 67.0 & 74.9 & 74.0 & 78.6 & 76.6 & 89.6 & 75.9 & 76.1 & 59.6 & 86.9 & 82.2 & 76.5 \\
Mistral-Large 2$^{\star}$ & 123B & 64.8 & 74.7 & 69.6 & 84.2 & 80.3 & 87.7 & 71.8 & 74.5 & 60.8 & 87.0 & 85.9 & 76.5 \\
gpt-4o-2024-05-13 & - & 68.1 & 76.8 & 71.4 & 79.8 & 78.5 & 86.5 & 69.0 & 77.5 & 59.6 & 83.6 & 84.3 & 75.9 \\
\rowcolor{gray!15}
HalluGuard-4B & 4B & 61.1 & 73.1 & \cellcolor{darkgreen}71.7 & 77.0 & \cellcolor{darkgreen}80.1 & 89.3 & \cellcolor{darkgreen}73.6 & 77.8 & \cellcolor{darkgreen}60.0 & \cellcolor{darkgreen}85.1 & \cellcolor{darkgreen}84.0 & \cellcolor{darkgreen}75.7 \\
Qwen2.5-72B-Instruct & 72B & 63.6 & 73.0 & 71.9 & 80.4 & 80.2 & 88.9 & 70.0 & 77.0 & 60.1 & 84.3 & 81.9 & 75.6 \\
Llama-3.1-70B-Instruct & 70B & 65.7 & 72.5 & 72.9 & 81.0 & 73.9 & 86.4 & 70.3 & 78.6 & 58.5 & 83.8 & 83.0 & 75.1 \\
Claude-3 Opus & - & 65.2 & 72.4 & 74.1 & 82.4 & 75.0 & 83.8 & 69.3 & 78.8 & 58.8 & 81.6 & 81.8 & 74.8 \\
Llama-3.3-70B-Instruct$^{\star}$ & 70B & 68.7 & 74.7 & 69.5 & 78.4 & 76.6 & 85.5 & 67.4 & 78.5 & 58.3 & 79.8 & 82.6 & 74.5 \\
Llama-3.1-405B-Instruct & 405B & 64.8 & 75.1 & 68.6 & 81.2 & 71.8 & 86.4 & 67.5 & 79.4 & 58.5 & 81.9 & 82.9 & 74.4 \\
gpt-4o-mini-2024-07-18 & - & 61.8 & 73.6 & 71.3 & 79.7 & 76.3 & 85.8 & 69.8 & 76.0 & 58.3 & 80.3 & 81.6 & 74.0 \\
\rowcolor{gray!15}
Qwen3-4B & 4B & \cellcolor{darkgreen}64.9 & \cellcolor{darkgreen}73.8 & 70.9 & \cellcolor{darkgreen}77.4 & 68.9 & \cellcolor{darkgreen}89.5 & 64.8 & \cellcolor{darkgreen}78.7 & 57.5 & 81.5 & 83.7 & 73.8 \\
Llama-3-70B-Instruct & 70B & 63.7 & 70.2 & 71.5 & 80.6 & 74.4 & 85.9 & 67.8 & 76.2 & 57.8 & 82.4 & 80.6 & 73.7 \\
Llama-3.1-8B-Instruct & 8B & 54.7 & 68.5 & 71.1 & 75.5 & 72.0 & 83.5 & 66.5 & 72.3 & 57.8 & 77.5 & 73.6 & 70.3 \\
Llama-3.2-1B-Instruct & 1B & 50.1 & 50.9 & 50.0 & 50.2 & 49.7 & 50.4 & 50.5 & 50.2 & 49.9 & 50.1 & 50.9 & 50.3 \\
Qwen3-0.6B$^{\star}$ & 0.6B & 20.5 & 43.4 & 15.9 & 26.2 & 26.5 & 81.1 & 23.6 & 69.4 & 38.5 & 25.5 & 14.9 & 35.0 \\
\bottomrule
\end{tabular}
}
\caption{\textbf{Evaluation on LLM-AggreFact.} Models are ordered by average balanced accuracy (BAcc Avg.; higher is better). HalluGuard-4B (ours), Qwen3-0.6B, 4B and 32B were evaluated using our specific prompt in \texttt{think} mode. All other results are taken from the public leaderboard. The higher score between HalluGuard-4B and Qwen3-4B is shaded in dark green. Alternating grey rows improve readability. $^{\star}$ Models used within our training pipeline.}
\label{tab:aggrefact_perf}
\end{table*}

\section{Experimental Setup}

\paragraph{Benchmark Dataset.} 
We evaluate on LLM-AggreFact~\cite{tang-etal-2024-minicheck}, a collection of human-annotated datasets designed to assess whether model-generated claims are supported by evidence documents. The benchmark spans diverse domains and incorporates real hallucinations from recent LLMs, directly aligning with our task of detecting if a claim is \texttt{grounded} or \texttt{hallucinated}. Importantly, it also includes RAGTruth~\cite{niu-etal-2024-ragtruth}, which is particularly relevant to our focus on hallucination mitigation in RAG (see Appendix~\ref{app:detailed_benchmark_datasets}). 


\paragraph{Evaluation Metric.} Performance is measured using balanced accuracy (BAcc)~\cite{brodersen2010balanced}, defined as
$\mathrm{BAcc} = \tfrac{1}{2}\left(\tfrac{\mathrm{TP}}{\mathrm{TP}+\mathrm{FN}} + \tfrac{\mathrm{TN}}{\mathrm{TN}+\mathrm{FP}}\right)$
where TP, TN, FP, and FN denote true positives, true negatives, false positives, and false negatives. We adopted BAcc to ensure comparability with prior work, as it was also used in the paper that introduced LLM-AggreFact.

\section{Results}

\paragraph{Evaluation on Benchmark.} As shown in Table~\ref{tab:aggrefact_perf}, \HalluGuard{} achieves an average BAcc of 75.7\%, improving upon its backbone Qwen3-4B (73.8) by +1.9 points, with strong gains on WiCE (+11.2) and ClaimVerify (+8.8). These scores are obtained in \texttt{think} mode using specific prompt and inference parameters (see Appendices~\ref{app:prompt_pairs_generation} and~\ref{app:inference_params}).

\HalluGuard{} is competitive with larger general-purpose LLMs (e.g., GPT-4o (75.9), Claude-3 Opus (74.8), Llama-3.3-70B (74.5), and Mistral-Large 2 (76.5). Compared to specialized models, HalluGuard-4B is behind Granite Guardian 3.3 (76.5) and MiniCheck-7B (77.4). However, these baselines are larger (8B and 7B parameters). HalluGuard-4B trails Granite Guardian by only 0.8 points and MiniCheck-7B by 1.7, while surpassing them on some benchmarks.

These results show that our fine-tuning pipeline transforms a lightweight backbone into a model that rivals both closed and open models, including general-purpose and specialized models, making \HalluGuard{} well suited for enterprise RAG applications where hallucination detection is crucial.

\paragraph{RAGTruth Detailed Evaluation.} 
This subset focuses on RAG settings, evaluating whether claims are supported by retrieved documents.

\HalluGuard{} achieves an average BAcc of 84.0\% like MiniCheck-7B (84.0) and surpasses Granite Guardian 3.3 (82.2) despite using roughly half their parameters. 
It correctly classifies 13,649 grounded claims and detects 984 hallucinations, missing only 282 (see Table~\ref{tab:ragtruth_conf}). This corresponds to a True Positive Rate (TPR) of 77.7\% and a True Negative Rate (TNR) of 90.7\%, showing that HalluGuard captures most hallucinations while preserving grounded content. This is essential for user trust. By combining high recall with evidence-grounded justifications, HalluGuard provides transparent decisions that users can verify, reinforcing its suitability in enterprise RAG applications.

\begin{table}[htbp]
\centering
\begin{tabular}{cc|cc}
\multicolumn{2}{c}{} & \multicolumn{2}{c}{\textbf{Predicted}} \\ 
\multicolumn{2}{c}{} & Hallucinated & Grounded \\ \hline
\multirow[c]{2}{*}{\rotatebox[origin=tr]{90}{\textbf{Actual}}}
& Hallucinated & 984 & 282  \\[1.5ex]
& Grounded & 1396 & 13649 \\ \hline
\end{tabular}
\caption{\textbf{Confusion Matrix on the RAGTruth Dataset.} Rows denote actual labels, columns denote predictions. The \texttt{hallucinated} label is treated as the positive class.}
\label{tab:ragtruth_conf}
\end{table}

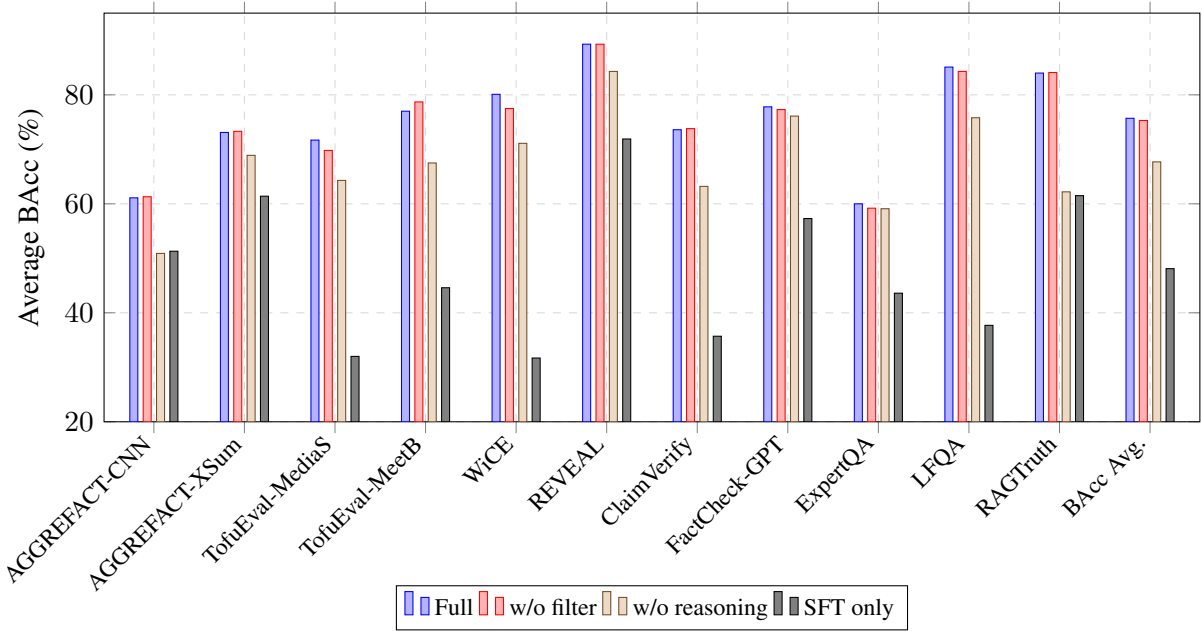
\begin{figure*}[h]
\centering
\begin{tikzpicture}
\begin{axis}[
    ybar,
    bar width=3pt,
    width=\textwidth,
    height=7cm,
    ylabel={Average BAcc (\%)},
    symbolic x coords={AGGREFACT-CNN,AGGREFACT-XSum,TofuEval-MediaS,TofuEval-MeetB,WiCE,REVEAL,ClaimVerify,FactCheck-GPT,ExpertQA,LFQA,RAGTruth, BAcc Avg.},
    xtick=data,
    xticklabel style={rotate=45, anchor=east, font=\small},
    ymin=20, ymax=95,
    legend style={at={(0.5,-0.4)}, anchor=north, legend columns=-1, font=\small},
    enlarge x limits=0.05,
    grid=both,
    grid style={dashed,gray!30},
]

\addplot coordinates {(AGGREFACT-CNN,61.1) (AGGREFACT-XSum,73.1) (TofuEval-MediaS,71.7) (TofuEval-MeetB,77.0) (WiCE,80.1) (REVEAL,89.3) (ClaimVerify,73.6) (FactCheck-GPT,77.8) (ExpertQA,60.0) (LFQA,85.1) (RAGTruth,84.0) (BAcc Avg.,75.7)};

\addplot coordinates {(AGGREFACT-CNN,61.3) (AGGREFACT-XSum,73.3) (TofuEval-MediaS,69.8) (TofuEval-MeetB,78.7) (WiCE,77.5) (REVEAL,89.3) (ClaimVerify,73.8) (FactCheck-GPT,77.3) (ExpertQA,59.2) (LFQA,84.3) (RAGTruth,84.1) (BAcc Avg.,75.3)};

\addplot coordinates {(AGGREFACT-CNN,50.9) (AGGREFACT-XSum,68.9) (TofuEval-MediaS,64.3) (TofuEval-MeetB,67.5) (WiCE,71.1) (REVEAL,84.3) (ClaimVerify,63.2) (FactCheck-GPT,76.1) (ExpertQA,59.1) (LFQA,75.8) (RAGTruth,62.2) (BAcc Avg.,67.7)};

\addplot coordinates {(AGGREFACT-CNN,51.3) (AGGREFACT-XSum,61.4) (TofuEval-MediaS,32.0) (TofuEval-MeetB,44.6) (WiCE,31.7) (REVEAL,71.9) (ClaimVerify,35.7) (FactCheck-GPT,57.3) (ExpertQA,43.6) (LFQA,37.7) (RAGTruth,61.5) (BAcc Avg.,48.1)};

\legend{Full, w/o filter, w/o reasoning, SFT only}
\end{axis}
\end{tikzpicture}
\caption{\textbf{Ablation of HalluGuard-4B.} Comparison of the full model and three variants on LLM-AggreFact.}
\label{fig:ablation}
\end{figure*}

\paragraph{Justification Evaluation.} A common baseline to evaluate generated text against reference is metrics such as ROUGE~\cite{lin_rouge_2004}. However, these metrics are inadequate for our task, because they cannot assess whether a justification is factually grounded in the source document and have been shown to correlate poorly with human judgments~\cite{wang_is_2023}. Thus, we adopt the G-Eval framework~\cite{liu_g-eval_2023}, using GPT-4o~\cite{openai2024gpt4technicalreport} as the evaluator. Concretely, for each RAGTruth document, we evaluate the justification of \PGL{}, \HalluGuard{}, and \PGS{}. G-Eval assesses four dimensions: Relevance, Consistency, and Coherence (each on a 1–5 scale), and Fluency (on a 1–3 scale).

As shown in Table~\ref{tab:g_eval_results}, the quality of the justification presents significant disparities. \QwenThirtyTwo{} achieves scores higher than \QwenZeroSix{} in all dimensions. Importantly, \HalluGuard{}, although almost an order of magnitude smaller than \QwenThirtyTwo{}, achieves comparable quality, indicating that ORPO effectively transfers strong model behavior to a smaller backbone.
In addition, fluency remains uniformly high, suggesting that the observed gains stem primarily from improved factual grounding and reasoning, rather than surface-level language quality.
Finally, these results show that \HalluGuard{} can match the quality of justification of a 32B parameter model.

\begin{table}[h]
\centering
\begin{tabular}{lcccc}
\toprule
\textbf{Model} & \textbf{Rel} & \textbf{Coh} & \textbf{Con} & \textbf{Flu} \\
\midrule
Qwen3-32B & 4.41 & 4.29 & 4.47 & 2.98 \\
HalluGuard-4B & 4.36 & 4.27 & 4.51 & 2.97 \\
Qwen3-0.6B & 3.72 & 3.65 & 3.58 & 2.75 \\
\bottomrule
\end{tabular}
\caption{\textbf{G-Eval Results.} Evaluation of justification on RAGTruth using four dimensions: Relevance (Rel), Coherence (Coh), Consistency (Con), and Fluency (Flu).}
\label{tab:g_eval_results}
\end{table}

\paragraph{Human Alignment Evaluation.} We evaluated the alignment of our preference construction based on heuristics (Section~\ref{ss:preference_training_dataset_construction}) with human judgments.

We sampled 100 preference tuples $z_i$, balancing grounded and hallucinated claims. Each tuple was assessed by two independent NLP expert annotators using the same criteria as those used to construct the preference dataset: correct classification, coherence of reasoning, and clarity of justification. 

Annotators were asked to indicate which completion they preferred between the two options. Importantly, they were blind to the labels and did not know which completion had been designated as \texttt{chosen} or \texttt{rejected} during dataset construction. To avoid bias, the completions were presented in random order and without any indications.

At the item level (75 pairs with full annotator agreement), \texttt{chosen} was preferred in 71 cases (94.7\%) vs.\ 4 for \texttt{rejected} ($p=3.4\times10^{-17}$, binomial test vs.\ 50\%, 95\% CI [0.89, 1.00]). At the annotation level (considering all 200 individual judgments), 83.5\% favored \texttt{chosen} ($p=4.7\times10^{-23}$, 95\% CI [0.79, 1.00]) (see Table~\ref{tab:human_alignment}).

\begin{table}[h]
\centering
\begin{tabular}{lc}
\toprule
\textbf{Evaluation level} & \textbf{Pref. for chosen} \\
\midrule
Item level ($n=75$) & 94.7\% \\
Annotation level ($n=200$) & 83.5\% \\
\bottomrule
\end{tabular}
\caption{\textbf{Human Alignment Results.} Annotators preferred the \texttt{chosen} completions (94.7\% of the 75 fully agreed items; 83.5\% of the 200 individual judgments).}
\label{tab:human_alignment}
\end{table}

These results show that our heuristic is closely aligned with human preferences. The annotators clearly favored \PGL{} over \PGS{}, both at the item level (94.7\%) and across all judgments (83.5\%), confirming that our heuristic provides an effective proxy for human preference.

\section{Ablation Study}

\paragraph{Impact of Consensus Filtering.} 
Applying LLM-based consensus filtering using independent evaluators (\IEone{} and \IEtwo{}) to preference tuples provides a small but decisive improvement. 
With filtering, HalluGuard reaches 75.7\% BAcc, compared to 75.3\% without it (–0.4\%). 
Although the gain is modest, it is crucial. In fact, without this component, HalluGuard falls behind Qwen2.5-72B-Instruct (75.6\%).

\paragraph{Contribution of Reasoning.} Disabling reasoning by using \texttt{/no\_think} in the prompt leads to a decrease in performance. In \texttt{think} mode, HalluGuard reaches a BAcc of 75.7\%, whereas in \texttt{non-think} mode the BAcc decreases to 67.6\% (–8.1\%). This represents the second largest drop in our ablation study, highlighting the critical role of reasoning in mitigating hallucinations.

\begin{figure}[H]
  \centering
  \includegraphics[width=0.90\linewidth]{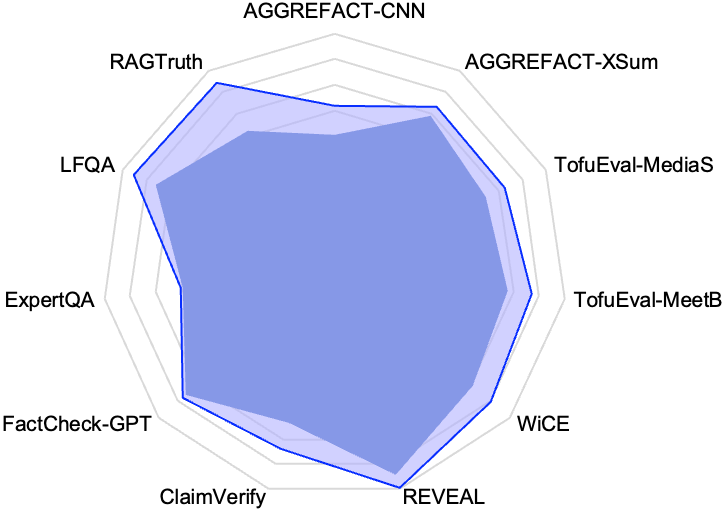}
  \caption{\textbf{Effect of Model Reasoning.} Radar plot comparing HalluGuard in \texttt{think} mode (lighter blue) vs.\ in \texttt{/no\_think} mode (darker blue).}
  \label{fig:ablation-reasoning}
\end{figure}

This is even more marked on RAGTruth, where reasoning improves BAcc (+21.8\%), with consistent gains across all other datasets (see Figure~\ref{fig:ablation-reasoning}).

\paragraph{Effect of Preference Alignment.} Replacing ORPO with SFT alone results in a the largest drop, with BAcc decreasing from 75.7\% to 48.1\% (–27.6\%). This indicates that preference alignment, as embedded in ORPO, plays a crucial role in enhancing the reliability and quality of reasoning.

\paragraph{Ablation Results.} Figure~\ref{fig:ablation} compares the full \HalluGuard{} model with three ablated variants on the benchmark datasets. The complete model consistently outperforms all variants, indicating that its robustness arises from the interaction of components rather than from any single factor. 

Consensus filtering yields a modest but consistent improvement of +0.4\% in BAcc, suggesting that pruning noisy preference pairs improves alignment. The second largest drop occurs when the reasoning traces are disable via \texttt{/no\_think} in the prompt, with BAcc decreasing by 8.1\% overall and reasoning providing a particularly large gain of +21.8\% on RAGTruth. Replacing ORPO with SFT alone further reduces performance by 27.6\%, confirming the importance of preference alignment. Together, these results support the retention of the entire pipeline to fine-tune \HalluGuard{}.

\section{Conclusion}
We presented HalluGuard, a 4B-parameter Small Reasoning Model designed to mitigate hallucinations in Retrieval-Augmented Generation while providing evidence-grounded justifications. 

Built on a domain-agnostic synthetic dataset with multi-stage curation and preference-based fine-tuning via ORPO and LoRA, we transform a compact backbone into a model that rivals or surpasses much larger LLMs, as well as recent specialized hallucination-detection models. 

In fact, \HalluGuard{} achieves competitive performance on LLM-AggreFact while providing justifications that are relevant, consistent, and comparable in quality to those of a 32B-parameter model.

Ablation studies also highlight the importance of reasoning traces, consensus filtering, and preference alignment in driving these gains. Thus, our findings demonstrate that carefully aligned small reasoning models can deliver both reliability and deployability for enterprise RAG applications, closing much of the gap with frontier LLMs.

To foster research, we will release HalluGuard and datasets under Apache 2.0 upon acceptance.

\section{Future Work}

In future work, we will (i) distinguish intrinsic and extrinsic hallucinations, and (ii) investigate multimodal extensions to support charts frequently present in enterprise documents. We will also release larger Qwen3-based variants (8B and 14B) to balance performance with deployment constraints.

\section*{Limitations}

\paragraph{Synthetic Data.} Although multiple filters are applied, synthetic claims may not fully capture the nuances of hallucinations encountered in real-world RAG applications, since the training is based on synthetic data. 

\paragraph{Output Formatting.}
To ensure deployment realism, we enforce a strict output structure: the response from \HalluGuard{} must be a JSON object containing \texttt{CLASSIFICATION} and \texttt{JUSTIFICATION} keys only. Any deviation from this is scored as incorrect and can underestimate performance.

\paragraph{Hallucination Coverage.} The current model merges intrinsic and extrinsic hallucinations under a single \texttt{hallucinated} label, which reduces explainability in settings where the distinction between different types of hallucination is important.

\paragraph{Language and Domain Generalization.} HalluGuard has been trained and evaluated on English data. Its performance in other languages or specialized domains remains uncertain.

\section*{Ethical Considerations}

As with any hallucination detection model, HalluGuard must be used with caution. Overflagging grounded claims may reduce user trust, while failing to detect hallucinations can lead to harmful errors further down the line. For this reason, HalluGuard should be used as a decision support tool rather than as a fully autonomous system, and should always be paired with human oversight. We therefore encourage responsible deployment in sensitive domains when integrating HalluGuard into real-world RAG applications.

\bibliography{halluguard_paper}

\appendix

\newpage

\section{Technical Reproducibility}
\label{app:reproducibility}

To facilitate reproducibility and transparency, we report the hardware and software environment used for all experiments. HalluGuard was fine-tuned on a single NVIDIA H100 PCIe GPU (80GB memory, TDP~350W) for 16 hours. Training consumed approximately 7.35 kWh of energy as calculated by the Machine Learning Impact Calculator (MLIC)~\cite{lacoste2019quantifying}. The experiments were carried out on a Linux server running CUDA~12.4.1 and PyTorch~2.4.0. We used the default random seeds and PyTorch settings.

\section{Prompt Template: Style Reformation}
\label{app:prompt_data_augmentation}

\begin{table}[h]
\centering
\small
\setlength{\tabcolsep}{4pt}
\renewcommand{\arraystretch}{1.1}
\begin{tabularx}{\linewidth}{lX}
\toprule
\textbf{Style} & \textbf{Instruction} \\
\midrule
\texttt{paraphrase} & Paraphrase the following text while retaining its original meaning. \\
\texttt{summarize} & Provide a concise summary of the following text. \\
\texttt{expand} & Expand on the following text by adding more details and context. \\
\texttt{news\_article} & Rewrite the following information as a news article. \\
\texttt{blog\_post} & Transform the following text into an engaging blog post. \\
\texttt{report} & Convert the following information into a formal report. \\
\texttt{story} & Rewrite the following text as a narrative story. \\
\texttt{dialogue} & Transform the following text into a dialogue between two characters. \\
\texttt{letter} & Rewrite the following text as a formal letter. \\
\texttt{social\_media\_post} & Transform the following text into a social media post. \\
\texttt{script} & Transform the following text into a script for a short video or play. \\
\texttt{interview} & Rewrite the following text as an interview between an interviewer and an expert. \\
\texttt{product\_description} & Transform the following text into a product description. \\
\texttt{review} & Rewrite the following text as a review of a product or service. \\
\texttt{news\_summary} & Summarize the following article into a concise news brief. \\
\texttt{formalize\_news} & Rewrite the following content in a formal journalistic style. \\
\texttt{meeting\_summary} & Rewrite the following text as if it were a summary of a team meeting. \\
\texttt{meeting\_dialogue} & Rewrite the following content as a conversation between multiple meeting participants. \\
\bottomrule
\end{tabularx}
\caption{Each style is applied to reform FineWeb raw data and increase stylistic diversity.}
\label{tab:prompts}
\end{table}

\section{Prompt Template: Claim Generation}
\label{app:prompt_claims_generation}

\paragraph{\texttt{grounded}}\mbox{}
\begin{lstlisting}[style=acljson,frame=single]
{
  "instructions": [
    "Generate a claim that is factually accurate and fully grounded in the provided context.",
    "Ensure that the claim is explicitly supported by the context - do not introduce information that is not directly verifiable from the context.",
    "Only return the claim as the answer. Do not include any additional text, explanation, or formatting."
  ],
  "context": <text>,
  "answer": ""
}
\end{lstlisting}

\paragraph{\texttt{hallucinated\_intrinsic}}\mbox{}
\begin{lstlisting}[style=acljson,frame=single]
{
  "instructions": [
    "Generate a claim that contradicts the provided context.",
    "The claim should remain fluent and grammatically correct but should be identifiable as incorrect upon a quick read.",
    "Only return the claim as the answer. Do not include any additional text, explanation, or formatting."
  ],
  "context": <text>,
  "answer": ""
}
\end{lstlisting}

\paragraph{\texttt{hallucinated\_extrinsic}}\mbox{}
\begin{lstlisting}[style=acljson,frame=single]
{
  "instructions": [
    "Generate a claim that includes information that cannot be verified within the provided context.",
    "Ensure the claim is plausible but requires external knowledge to verify its accuracy.",
    "Only return the claim as the answer. Do not include any additional text, explanation, or formatting."
  ],
  "context": <text>,
  "answer": ""
}
\end{lstlisting}

\section{Prompt Template: Synthetic Pairs}
\label{app:prompt_pairs_generation}

\begin{lstlisting}[style=acljson,frame=single]
{
  "instructions": [
    "You will be given a document and a claim. Determine whether the claim is 'GROUNDED' or 'HALLUCINATED' based on the document.",
    "A 'GROUNDED' claim is factually accurate and fully supported by the information provided in the document. It should be directly verifiable from the document.",
    "A 'HALLUCINATED' claim is either:",
    "  - Intrinsically incorrect: It contradicts the information provided in the document, or",
    "  - Extrinsically incorrect: It includes information that cannot be verified within the document and requires external knowledge to assess its accuracy.",
    "Return the classification as the answer (i.e., GROUNDED or HALLUCINATED). Include justification."
  ],
  "document": <document>,
  "claim": <claim>,
  "answer": { "CLASSIFICATION": "", "JUSTIFICATION": "" }
}
\end{lstlisting}

\section{Prompt Template: Consensus Filter}
\label{app:prompt_filtering_generation}

\begin{lstlisting}[style=acljson,frame=single]
{
  "instructions": [
    "You will be given a document and a claim, along with two responses (RESPONSE_A and RESPONSE_B).",
    "Determine which response is better based on classification correctness, thinking coherence and clarity, and justification quality.",
    "Return your answer as either 'RESPONSE_A' or 'RESPONSE_B', without any justification."
  ],
  "examples": <examples>,
  "document": <document>,
  "claim": <claim>,
  "RESPONSE_A": <response_a>,
  "RESPONSE_B": <response_b>,
  "best_response": ""
}
\end{lstlisting}

\section{Benchmark Datasets}
\label{app:detailed_benchmark_datasets}

LLM-AggreFact includes the following datasets: AGGREFACT~\cite{tang-etal-2023-understanding}, a factual consistency benchmark for summarization; TofuEval~\cite{tang-etal-2024-tofueval}, a dialogue summarization benchmark with LLM summaries annotated for factual consistency; WiCE~\cite{kamoi-etal-2023-wice}, a textual entailment dataset of Wikipedia claims and cited sources; REVEAL~\cite{jacovi2024chainofthoughtstrongweakestlink}, which evaluates reasoning chains in open-domain QA with sentence-level attribution labels against retrieved Wikipedia passages; ClaimVerify~\cite{liu-etal-2023-evaluating}, which assesses generative search engine responses by verifying check-worthy sentences against cited documents with binary factuality labels; FactCheck-GPT~\cite{wang2024factcheckbenchfinegrainedevaluationbenchmark}, which decomposes LLM responses to search queries into atomic facts; ExpertQA~\cite{malaviya-etal-2024-expertqa}, consisting of expert-curated queries across 32 domains where system responses are verified against evidence documents; LFQA~\cite{chen2023understandingretrievalaugmentationlongform}, where LLM long-form answers conditioned on retrieved or random documents are labeled; and RAGTruth~\cite{niu-etal-2024-ragtruth}, a retrieval-augmented generation benchmark where outputs grounded in retrieved passages are annotated. 

\section{Reward Gap Across Training Epochs}
\label{app:epoch_reward_margins}
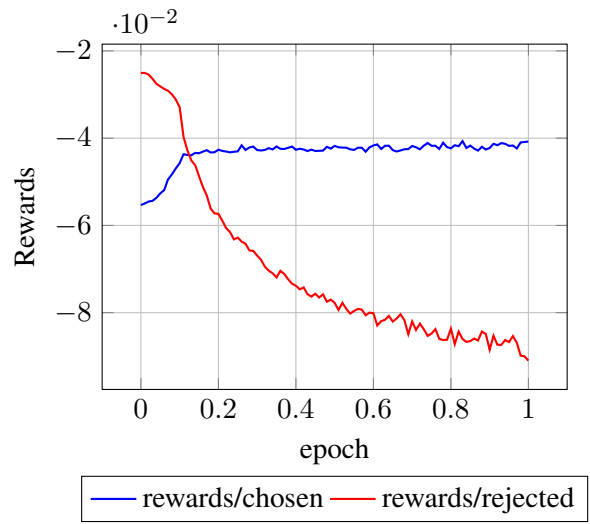
\begin{figure}[htbp]
\centering
\begin{tikzpicture}
\begin{axis}[
  xlabel={epoch},
  ylabel={Rewards},
  grid=both,
  legend pos=north west,
  legend cell align=left,
  width=\columnwidth,
  height=0.8\columnwidth,
  legend style={
  at={(0.5,-0.25)},
  anchor=north,
  legend columns=2}
]

\addplot+[mark=none, thick] coordinates {
(0,-0.0553377)
(0.01,-0.0549627)
(0.02,-0.0545252)
(0.03,-0.0543622)
(0.04,-0.0536907)
(0.05,-0.0526539)
(0.06,-0.0519124)
(0.07,-0.0494586)
(0.08,-0.0482898)
(0.09,-0.0469407)
(0.1,-0.0457255)
(0.11,-0.0436826)
(0.12,-0.0438574)
(0.13,-0.0439391)
(0.14,-0.0433851)
(0.15,-0.0434607)
(0.16,-0.0430833)
(0.17,-0.0427538)
(0.18,-0.043285)
(0.19,-0.0432274)
(0.2,-0.0426557)
(0.21,-0.0429261)
(0.22,-0.0430679)
(0.23,-0.0432585)
(0.24,-0.0431183)
(0.25,-0.0430312)
(0.26,-0.0416458)
(0.27,-0.042715)
(0.28,-0.0421367)
(0.29,-0.041936)
(0.3,-0.0427325)
(0.31,-0.0428281)
(0.32,-0.0426966)
(0.33,-0.0422864)
(0.34,-0.0425919)
(0.35,-0.041912)
(0.36,-0.0424427)
(0.37,-0.0424825)
(0.38,-0.0422157)
(0.39,-0.041918)
(0.4,-0.0426696)
(0.41,-0.0424277)
(0.42,-0.0426081)
(0.43,-0.042961)
(0.44,-0.0426281)
(0.45,-0.0429391)
(0.46,-0.0428995)
(0.47,-0.0428555)
(0.48,-0.0419838)
(0.49,-0.0424071)
(0.5,-0.0417745)
(0.51,-0.0420937)
(0.52,-0.0421631)
(0.53,-0.0421781)
(0.54,-0.0425568)
(0.55,-0.0426976)
(0.56,-0.0421947)
(0.57,-0.0421968)
(0.58,-0.0431017)
(0.59,-0.0421078)
(0.6,-0.0416776)
(0.61,-0.0414356)
(0.62,-0.0424873)
(0.63,-0.0417467)
(0.64,-0.0417317)
(0.65,-0.0427466)
(0.66,-0.0430521)
(0.67,-0.04282)
(0.68,-0.0425427)
(0.69,-0.042508)
(0.7,-0.0417905)
(0.71,-0.0420949)
(0.72,-0.0425466)
(0.73,-0.0417201)
(0.74,-0.0411139)
(0.75,-0.0417436)
(0.76,-0.0417543)
(0.77,-0.0424713)
(0.78,-0.0410401)
(0.79,-0.0420671)
(0.8,-0.0423239)
(0.81,-0.0416273)
(0.82,-0.0418375)
(0.83,-0.0406877)
(0.84,-0.0422618)
(0.85,-0.0417085)
(0.86,-0.0423964)
(0.87,-0.0428563)
(0.88,-0.0419727)
(0.89,-0.0426029)
(0.9,-0.0422554)
(0.91,-0.0412888)
(0.92,-0.0416197)
(0.93,-0.0411275)
(0.94,-0.0412962)
(0.95,-0.0417696)
(0.96,-0.0416851)
(0.97,-0.0423631)
(0.98,-0.0409801)
(0.99,-0.0408801)
(1,-0.0407801)
};
\addlegendentry{rewards/chosen}

\addplot+[mark=none, thick] coordinates {
(0,-0.025066)
(0.01,-0.0250477)
(0.02,-0.0254336)
(0.03,-0.0264063)
(0.04,-0.0275453)
(0.05,-0.0281382)
(0.06,-0.0287067)
(0.07,-0.0291373)
(0.08,-0.0299565)
(0.09,-0.0311518)
(0.1,-0.0328953)
(0.11,-0.0397154)
(0.12,-0.0427859)
(0.13,-0.0451017)
(0.14,-0.0461798)
(0.15,-0.0487137)
(0.16,-0.051192)
(0.17,-0.0531374)
(0.18,-0.0561552)
(0.19,-0.0572445)
(0.2,-0.0573275)
(0.21,-0.0589279)
(0.22,-0.0605951)
(0.23,-0.0615337)
(0.24,-0.0631908)
(0.25,-0.0628058)
(0.26,-0.0637065)
(0.27,-0.0642119)
(0.28,-0.065727)
(0.29,-0.065848)
(0.3,-0.0669293)
(0.31,-0.0678777)
(0.32,-0.0694688)
(0.33,-0.0703888)
(0.34,-0.0709734)
(0.35,-0.0719266)
(0.36,-0.0704186)
(0.37,-0.0711417)
(0.38,-0.0723493)
(0.39,-0.0733557)
(0.4,-0.0738279)
(0.41,-0.0746442)
(0.42,-0.0742145)
(0.43,-0.0757772)
(0.44,-0.0763166)
(0.45,-0.0756507)
(0.46,-0.0765289)
(0.47,-0.0757951)
(0.48,-0.077508)
(0.49,-0.0770047)
(0.5,-0.077757)
(0.51,-0.0793714)
(0.52,-0.0777462)
(0.53,-0.0791305)
(0.54,-0.0802085)
(0.55,-0.0796055)
(0.56,-0.0791451)
(0.57,-0.0792781)
(0.58,-0.0806044)
(0.59,-0.0799759)
(0.6,-0.0801296)
(0.61,-0.0829359)
(0.62,-0.0819429)
(0.63,-0.0816242)
(0.64,-0.0807409)
(0.65,-0.0820379)
(0.66,-0.0813284)
(0.67,-0.0803612)
(0.68,-0.081744)
(0.69,-0.0849237)
(0.7,-0.0820212)
(0.71,-0.0839741)
(0.72,-0.0825202)
(0.73,-0.0837972)
(0.74,-0.0853249)
(0.75,-0.0847978)
(0.76,-0.0837718)
(0.77,-0.0860062)
(0.78,-0.0862737)
(0.79,-0.0862154)
(0.8,-0.0836654)
(0.81,-0.087204)
(0.82,-0.0843342)
(0.83,-0.0860373)
(0.84,-0.0866982)
(0.85,-0.0865179)
(0.86,-0.0859302)
(0.87,-0.0863691)
(0.88,-0.0843316)
(0.89,-0.0848712)
(0.9,-0.0884845)
(0.91,-0.085298)
(0.92,-0.0873602)
(0.93,-0.0873738)
(0.94,-0.0862717)
(0.95,-0.0867469)
(0.96,-0.0853576)
(0.97,-0.0868537)
(0.98,-0.0898607)
(0.99,-0.09)
(1,-0.091)
};
\addlegendentry{rewards/rejected}

\end{axis}
\end{tikzpicture}
\caption{The gap between \texttt{chosen} and \texttt{rejected} responses increases over training, showing that the model progressively learns to prefer \texttt{chosen} examples while assigning lower rewards to \texttt{rejected} ones.}
\label{fig:epoch-rewards}
\end{figure}

\section{Inference Parameters}
\label{app:inference_params}

\begin{table}[h]
\centering
\begin{tabular}{p{0.25\linewidth} p{0.35\linewidth} p{0.2\linewidth}}
\toprule
\textbf{Parameter} & \textbf{Non-Thinking} & \textbf{Thinking} \\
\midrule
temperature & 0.7 & 0.6 \\
min\_p & 0.0 & 0.0 \\
top\_p & 0.8 & 0.95 \\
top\_k & 20 & 20 \\
\bottomrule
\end{tabular}
\caption{Inference parameters used in our experiments, following the recommended Qwen settings for non-thinking and thinking modes.}
\label{tab:halluguard_inference_params}
\end{table}

\newpage

\section{Fine-Tuning Configuration}
\label{app:fine_tuning_protocol}

\begin{table}[h]
\centering
\begin{tabular}{p{0.4\linewidth} p{0.45\linewidth}}
\toprule
\textbf{Parameter} & \textbf{Value} \\
\midrule
lora\_layers\_attn & \texttt{q\_proj, k\_proj, v\_proj} \\
lora\_layers\_ffn  & \texttt{gate\_proj, up\_proj, down\_proj} \\
lora\_rank & 16 \\
lora\_alpha & 16 \\
lora\_dropout & 0 \\
precision & \texttt{bfloat16} \\
epochs & 1 \\
batch\_size & 2 \\
grad\_accumulation & 4 \\
effective\_batch\_size & 8 \\
optimizer & AdamW (8-bit) \\
learning\_rate & $1\times10^{-6}$ \\
lr\_schedule & linear \\
orpo\_beta & 0.1 \\
max\_seq\_len & 32768 \\
\bottomrule
\end{tabular}
\caption{The setup trains $\sim$33M parameters (0.81\% of the full model) using LoRA for 1 epoch.}
\label{tab:finetune_config}
\end{table}

\section{ORPO Preference Tuple: Full Example}
\label{app:preference_data}

\begin{lstlisting}[style=acljson, frame=single]
{
  "prompt": "{
    "instructions": [
      "You will be given a document and a claim. Determine whether the claim is 'GROUNDED' 
      or 'HALLUCINATED' based on the document.",
      "A 'GROUNDED' claim is factually accurate and fully supported by the information provided in the document. It should be directly verifiable from the document.",
      "A 'HALLUCINATED' claim is either:",
        "  - Intrinsically incorrect: It contradicts the information provided in the document, or",
        "  - Extrinsically incorrect: It includes information that cannot be verified within the document and requires external knowledge to assess its accuracy.",
    "Return the classification as the answer (i.e., GROUNDED or HALLUCINATED). Include justification."
    ],
    "document": "'As I stepped into the charming St. Blasius parish... (truncated) ...'",
    "claim": "'St. Blasius Church is over 850 years old and was built during... (truncated) ...'",
    "answer": { "CLASSIFICATION": "", "JUSTIFICATION": "" }
  }",
  "chosen": [
    {
      "role": "assistant",
      "content": "<think>Okay, let's tackle this. The user wants to determine if the claim about St.    Blasius Church... (truncated) ... Therefore, the claim is GROUNDED</think>
      {
      "CLASSIFICATION": "GROUNDED",
      "JUSTIFICATION": "The claim that 'St. Blasius Church is over 850 years old and was built 
      during the reign of King Stephen' is directly supported by the document... (truncated) ..."
      }"
    }
  ], 
  "rejected": [
    {
      "role": "assistant",
      "content": "<think>Okay, let's tackle this. The user wants me to determine if the claim... (truncated) ... Since the document contradicts the claim, the claim is intrinsically hallucinated.</think>
    {
    "CLASSIFICATION": "HALLUCINATED",
    "JUSTIFICATION": "The document explicitly states: 'The church itself was over 850 years old, built during the reign of King Stephen ... (truncated) ...'"
    }"
    }
  ]
}
\end{lstlisting}

\twocolumn

\end{document}